\documentclass[10pt,twocolumn,letterpaper]{article}

\usepackage[pagenumbers]{cvpr} 

\usepackage{graphicx}
\usepackage{amsmath}
\usepackage{amssymb}
\usepackage{booktabs}
\usepackage{bbding}

\usepackage[pagebackref,breaklinks,colorlinks]{hyperref}

\usepackage[capitalize]{cleveref}
\crefname{section}{Sec.}{Secs.}
\Crefname{section}{Section}{Sections}
\Crefname{table}{Table}{Tables}
\crefname{table}{Tab.}{Tabs.}

\def\cvprPaperID{54} 
\def\confName{CVPR}
\def\confYear{2023}

\begin{document}

\title{Multimodal Feature Extraction and Fusion for Emotional Reaction Intensity Estimation and Expression Classification in Videos with Transformers}  

\author{Jia Li\thanks{Equal contribution}, 
Yin Chen\footnotemark[1],
Xuesong Zhang, 
Jiantao Nie, 
Ziqiang Li,
Yangchen Yu,
Yan Zhang\footnotemark[4], \\
Richang Hong\thanks{Corresponding author},
Meng Wang\footnotemark[2]
\\
Hefei University of Technology, China\\
\footnotemark[4] Anhui University, China \\
{\tt\small jiali@hfut.edu.cn} \\
{\tt\small \{chenyin,xszhang\_hfut,2022111086,2019212292\}@mail.hfut.edu.cn}\\
{\tt\small zhangyan@ahu.edu.cn} \\
{\tt\small \{luoyixuan1029,eric.mengwang,hongrc.hfut\}@gmail.com}
}


\maketitle

\begin{abstract}
In this paper, we present our advanced solutions to the two sub-challenges of Affective Behavior Analysis in the wild (ABAW) 2023: the Emotional Reaction Intensity (ERI) Estimation Challenge and Expression (Expr) Classification Challenge. ABAW 2023 aims to tackle the challenge of affective behavior analysis in natural contexts, with the ultimate goal of creating intelligent machines and robots that possess the ability to comprehend human emotions, feelings, and behaviors. For the Expression Classification Challenge, we propose a streamlined approach that handles the challenges of classification effectively. However, our main contribution lies in our use of diverse models and tools to extract multimodal features such as audio and video cues from the Hume-Reaction dataset. By studying, analyzing, and combining these features, we significantly enhance the model's accuracy for sentiment prediction in a multimodal context. Furthermore, our method achieves outstanding results on the Emotional Reaction Intensity (ERI) Estimation Challenge, surpassing the baseline method by an impressive 84\% increase, as measured by the Pearson Coefficient, on the validation dataset.
\end{abstract}

\section{Introduction}
\label{sec:intro}
Emotions are psychological states that arise in human beings when stimulated by the external environment, and they are able to reflect a person's current physiological and psychological state, making an impact on people's perception of things and future decisions. Emotions are often accompanied by distinct physiological characteristics, which can be analyzed to identify them. In this paper, we present in detail our solution to the Expression Classification Challenge and Emotional Reaction Intensity (ERI) Estimation Challenge.

The Expression Classification Challenge dataset comprises 546 videos that contain annotations for the six basic expressions (i.e., anger, disgust, fear, happiness, sadness, surprise), along with a neutral state and a category for other expressions/affective states. This challenge requires outputting the expression type for each frame.

In the Emotional Reaction Intensity (ERI) Estimation Challenge,  Hume-Reaction dataset was used. Each participant subjectively annotated his or her intensity of 7 emotional experiences\cite{christ2022muse}, including adoration, amusement, anxiety, disgust, empathic pain, fear, and surprise. 
The model is required to estimate the intensity of the participants' emotional experiences caused by watching the video. The model hence needs to analyze the visual and audio information present in the video and make accurate predictions about the emotional state of the participants. The task is challenging because the model needs to correctly identify the emotions that the participants are experiencing and measure their intensity while taking into account the context of the video.
Participants tend to express their emotions through their voices, body movements, and expressions, and it is through these factors that multimodal emotion recognition can identify and predict emotions. 

Traditional emotion recognition is based on a single modality, however, when a single factor is affected, for example, when facial expressions are not clear due to sunlight or when voices are masked by ambient sounds, the effectiveness of emotion prediction is significantly reduced. To further augment its feature representation capacity, 
We first pre-train the model on the AffectNet7 dataset\cite{mollahosseini2017affectnet} using the Posterv2\cite{mao2023poster} and ViT\cite{dosovitskiy2020image} models, respectively. Then, we combine these pre-trained models to obtain a novel pretrained PosterV2-ViT model that is specifically designed for extracting features on the Hume-Reaction dataset. This approach significantly improves the performance of model and enables us to achieve state-of-the-art results in this Challenge.
Furthermore, to fully harness the potential of audio and video features, we incorporated a modal interaction module to effectively extract and capture the common relationships between these two modalities. 
Our contributions in enhancing feature extraction ability and modality interaction have significant implications for emotion recognition in unstable environments with varied stimuli.

    Our contributions can be summarized as follows.
    
1. We propose a cross modality interaction approach to combine audio and visual features. Specifically, we extract visual features using a novel PosterV2-ViT model, which significantly improves the performance of our mmodel.

2.  For the emotional reaction intensity estimation challenge, our method achieves a Mean Person score of 0.4394 on the Hume-Reaction validation set. The experiment result also suggests that our method is effective in capturing the inter-person variations in emotional reactions.

3. For the expression classification challenge, our simple approach achieves 0.346 for the F1 score on the Aff-Wild2 validation set. 


\label{sec:meth}
\begin{figure*}[t]
  \centering
   \includegraphics[width=0.85\linewidth]{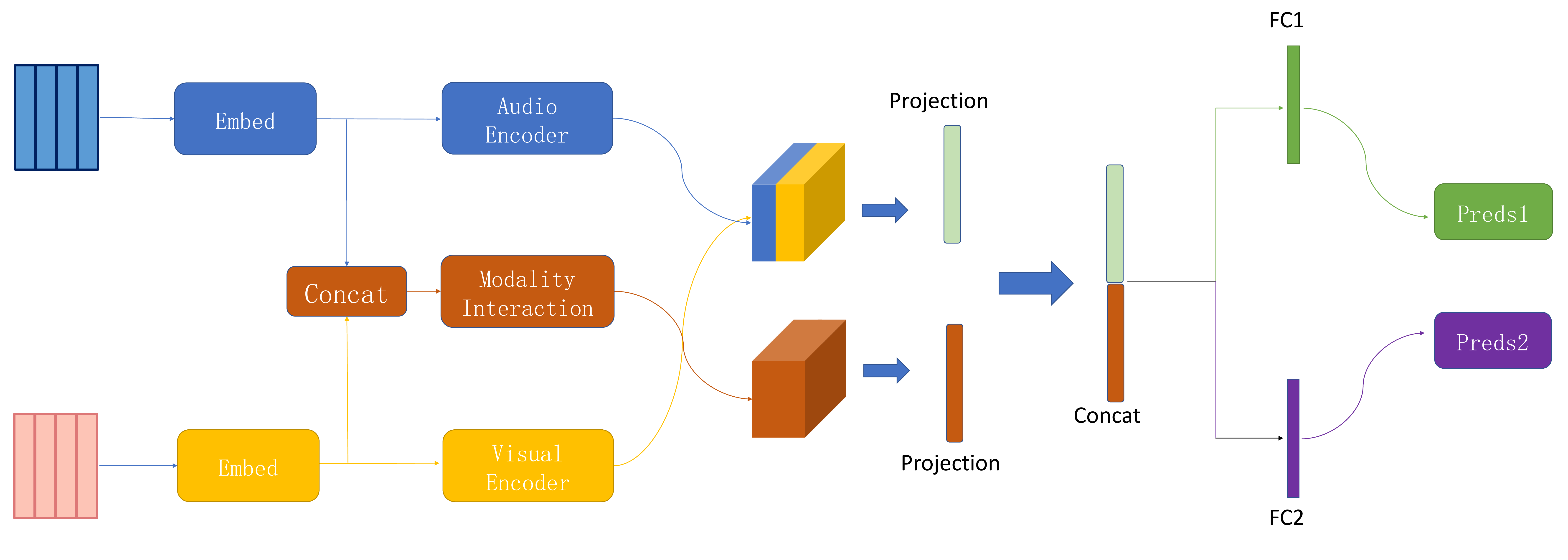}
   \caption{The framework of our network architecture. It consists of three main branches, i.e., an audio feature encoding module, a visual feature encoding module, and an audio-visual modality interaction module. Each module comprises a single-layer transformer encoder.}
   \label{fig:fram}
\end{figure*}
\section{Related Work}
\label{sec:rel}
Human emotions can be realistically reflected by facial
expressions in most cases, so recognizing human expressions and estimating emotional reaction intensity are crucial for human-computer interaction (HCI) systems\cite{kollias2022abaw2,kollias2021distribution,kollias2021analysing,kollias2021affect,kollias2020analysing,kollias2019face,kollias2023abaw4,zafeiriou2017aff}. In this section, we will introduce some related works about emotion recognition and emotional reaction intensity.

\subsection{Emotional Reaction Intensity}
Emotion recognition research has primarily focused on the classification of basic emotions, including positive, negative, and neutral states. However, there have been recent efforts\cite{kosti2017emotion,khattak2022efficient,siddiqui2022robust} to differentiate more nuanced emotional categories such as anger, happiness, and sadness.

Building upon emotion recognition, some studies attempt to estimate the intensity or degree of emotional reactions. Emotional Reaction Intensity Estimation Challenge were presented in MuSe 2022\cite{amiriparian2022muse} and ABAW5\cite{kollias2023abaw}. The objective of this challenge was to assess the magnitude of response elicited by seven distinct expression types, encompassing adoration, amusement, anxiety, disgust, empathic-pain, fear and surprise, with intensity values between 0 to 100.

The TEMMA model proposed by Li\cite{li2022hybrid} at Muse 2022 won the championship in this challenge. Resnet-18 feature and DeepSpectrum feature were used in \cite{li2022hybrid}. 
The researchers\cite{li2022hybrid} use a Resnet-18 model to extract static visual features from the videos, and a DeepSpectrum\cite{amiriparian2017snore-deepspectrum} model to extract audio features from the videos. These features are then input into the model as training data. Among them, the DeepSpectrum model is a spectral analysis algorithm that extracts audio features from the videos using the frequency domain. By combining the visual and audio features, the model is able to make more accurate predictions about the expression type and intensity of the participants' emotional experiences.

In addition, there were many exceptional contributions to the competition that emerged, and the methods used by these teams also achieved remarkable results. FaceRNET\cite{kollias2023facernet} is a convolutional recurrent neural network with a routing mechanism atop, which yield excellent results on the Hume Reaction dataset. They\cite{kollias2023facernet} used a representation extractor network to extract emotion descriptors, and a RNN network to capture
temporal information with a mask layer. Former-DFER+MLGCN\cite{wang2022emotional}  also proposes an end-to-end network framework that has achieved remarkable performance on the Hume-Reaction dataset. To better utilize multimodal information, ViPER\cite{vaiani2022viper} proposed a multimodal architecture for emotion recognition which achieves excellent performance on the ERI task as well.

\subsection{Expression Classification}
In the field of facial expression recognition, numerous datasets and research methods have been proposed. In the term of dataset, VGGFace\cite{parkhi2015deep} and CASIA-WebFace that are widely used for training and evaluating deep learning models in face recognition. The Expression Classification Challenge is based on the Aff-Wild2 dataset which is a naturalistic affective behavior analysis dataset. It aims to provide high-quality facial data for expression analysis. 
As for the term of research methods, the champion in expression classification challenge is dedicate in Multimodal feature extraction in the ABAW3 Competition. They make static vision feature extractor pre-trained and integrate the static and dynamic multimodal feature\cite{1zhang2022transformer}. As the winner of expression classification challenge in ABAW5, Zhang et al\cite{zhang2023facial} introduced a two-branch collaboration training strategy that randomly interpolates the logits space.

\section{Methodology}
\subsection{Emotional Reaction Intensity Estimation}
\subsubsection{Model Structure}
Figure \ref{fig:fram} illustrates the overall framework of our network architecture, which consists of three main branches: an audio feature encoding module, a visual feature encoding module, and an audio-visual modality interaction module. Each module comprises a single-layer transformer encoder. 

Firstly, the extracted audio and visual features were passed through a embedding layer(Figure.\ref{fig:fram} Embed Box) to unify the dimensions of the input. The embedding layer we used is 1-dimensional convolution.
\begin{equation}
    h^d_m = Embed(input_m), m \in \{a,v\}
    \label{eq:embed}
\end{equation}
The function $Embed$ in Equation.\ref{eq:embed} means the embedding layer, variable $input_m$ denotes the input features, where $m$ represents different modalities, $a$ refers to audio, $v$ refers to visual. $h$ refers to the embedded features, and $d$ represents the dimension of the output features. 

Next, we passed the extracted audio and visual features through separate feature encoders to extract further features. Additionally, we concatenated the audio and visual features and passed them through a modal interaction encoder to extract features between the two modalities:
\begin{equation}
    att = MHA(x)
    \label{eq:encoder}
\end{equation}
\begin{equation}
    ouput = concat(x,att,trans(x))
    \label{eq:encoder2}
\end{equation}
Equation \ref{eq:encoder}, \ref{eq:encoder2} define the processing procedure of the Encoder in Figure. \ref{fig:fram}, while $x$ is the input vector, $MHA$ is Multi-Head Attention\cite{vaswani2017attention} and trans is the transformer\cite{vaswani2017attention} encoder layer.
\begin{equation}
    g^{3d}_m= Encoder_m(h^d_m), m \in \{a,v\} 
    \label{eq:encoderav}
\end{equation}
\begin{equation}
    h^{2d}_{av} = concat(h_a,h_v) 
    \label{eq:encodercat}
\end{equation}
\begin{equation}
    g^{6d}_{av} = Encoder_{av}(h^{2d}_{av})
    \label{eq:encoderv}
\end{equation} 
Here, $Encoder_m$ represents the modality encoder, and 
$g_m$ represents the output features of the modality encoder. We first concatenate a,v features then encode them with modality interaction encoder. This procedure is defined in equation \ref{eq:encoderav},\ref{eq:encodercat},\ref{eq:encoderv}. 

Afterwards, we concatenated the audio and visual modality encoder features to create $g_{cat}$.
\begin{equation}
    g^{6d}_{cat} = concat(g^{3d}_a,g^{3d}_v)
    \label{eq:concat}
\end{equation}

Finally, we projected the modality features $g_{av}$ and $g_{cat}$
 to a lower dimension using a projection layer, and concatenated the output features before passing them through a fully connection layer separately to obtain the final results.

\subsubsection{Feature Extraction}
In our solution for ERI, we utilized DeepSpectrum\cite{amiriparian2017snore-deepspectrum} audio features and PosterV2-Vit visual features as our input features, both of which yielded excellent results. 

DeepSpectrum\cite{amiriparian2017snore-deepspectrum} refers to a 1024 dimensional feature extracted from speech signals by deep neural networks. We utilized DenseNet121\cite{huang2017densely}, trained on ImageNet\cite{russakovsky2015imagenet}, for feature extraction of DeepSpectrum. The visual features were extracted using the pretrained model PosterV2-Vit on the AffectNet7 dataset, resulting in a feature vector of 1536 dimensions. The PosterV2-Vit model was obtained from the joint training of the PosterV2\cite{mao2023poster} and ViT models after individually pretraining on the AffectNet7 dataset. The final model achieved an accuracy of 67.57\% on the AffectNet7 test set.

\subsubsection{Loss Functions}
Estimating Emotional Reaction Intensity is a multi-regression task. However, based on observations from our data annotation, it was found that each label consistently has a value of 1 for intensity. Therefore, we decoupled this intensity estimation task into two separate tasks: one for multi-regression and the other for classification. As a result, we utilized two different loss functions: mean squared error (MSE) loss and cross entropy loss. 
\begin{equation}
    Loss_{ERI} = \alpha L_{reg} + \beta L_{class}
    \label{eq:loss}
\end{equation}
As shown in Equation \ref{eq:loss}, the loss function for regression tasks using MSE loss, and for classification tasks using cross entropy loss, can be denoted as $L_{reg}$ and $L_{class}$, respectively. $\alpha$ and $\beta$ are the loss weight for $L_{reg}$ and $L_{class}$.

\subsection{Expression Classification}
\label{sec:exprcls}
Expression classification and emotional reaction intensity estimation are two very different tasks that are based on different datasets. Therefore, we propose a new but simple method to handle the expression classification task.

To enhance the robustness of the model on the expression classification challenge, we first obtain different pre-trained models on different face datasets.
Specifically, we pre-trained the InceptionResnetV1 model\cite{szegedy2017inception} on the casia-webface\cite{webyi2014learning} and VGGfacev2 datasets\cite{parkhi2015deepvgg}, respectively. We use the Regnet model\cite{regradosavovic2020designing,5phan2022facial} pre-trained on the imageNet dataset to improve the generalization ability of the model\cite{deng2009imagenet}.

Transformed-based methods are usually effective in capturing attention information of modality\cite{vaswani2017attention}. Therefore, we stitch these features and then feed them uniformly into a transformed-based feature fusion module for training. 


Deng et al.\cite{merdeng2020multitask} showed that there is a general category imbalance in the dataset. Focal loss rebuilding the standard cross entropy loss to solve class imbalance, which can be defined as :
\begin{equation}
    Loss_{Expr} = -\alpha_{t}(1-p_{t})^{\gamma}log(p_t) 
    \label{eq:loss}
\end{equation}
where parameter $\alpha_t \geqq 0$ and $\gamma \in [0, 5]$

\section{Experiments}
\label{sec:expr}
\subsection{Datasets}
Hume-Reaction\cite{christ2022muse} dataset: This multi-modal dataset comprises approximately 75 hours of video recordings, captured via a webcam located in the participants' homes. The dataset consists of recordings of 2222 participants belonging to two different cultures, South Africa and the United States. Each sample within the dataset has been self-annotated by the participants, indicating the intensity of 7 emotional experiences, including Adoration, Amusement, Anxiety, Disgust, Empathic Pain, Fear, and Surprise, on a scale from 1 to 100.

Aff-Wild2 database \cite{kollias2019expression}: For the Expression Classification Challenge, the Aff-Wild2 database will be used. About 2.7M frames in this database are annotated in terms of the 6 basic expressions (i.e., anger, disgust, fear, happiness, sadness, surprise), plus the neutral state and a category 'other' that denotes expressions states other than the 6 basic ones. 

\subsection{Evaluation Metrics}

For the expression classification challenge, we calculate the F1-Score for every class to assess the prediction results. With regards to the intensity estimation of emotional reactions, we employ Pearson’s Correlation Coefficient as the metric for each class.

\subsection{Experimental Setup}
Throughout the experiment, we adopt the Adam optimizer with a fixed learning rate, which is 1e-4. Additionally, we employed an early stopping technique, which stopped the training process when the metric failed to increase for 15 consecutive epochs. The evaluation metric being used is Pearson's correlation coefficient (p). The $\alpha$ and $\beta$ in equation \ref{eq:loss} was set 1.0 and 0.01, respectively. Moreover, we used Exponential Moving Average to improve the model's generalization ability, with a decay rate of 0.99. To prevent the gradient explosion issue that could negatively impact the model optimization, we applied gradient clipping with max\_norm=0.1 and norm\_type=2.

In the experiment, we set the dimension of the embedding to 256, the depth of the modality encoder to 1, the number of attention heads to 2, and the dropout to 0.5. Moreover, the depth of the modality interaction module was set to 1, the number of attention heads to 8, and the dropout to 0.1.
During the training, the batch size is set to 256, and trained for 100 epochs. 

For some experimental settings for expression classification challenge, please refer to the baseline\cite{5phan2022facial} we adopted. We used Pytorch framework to conducted all experiments on a NVIDIA RTX 3090 GPU.

\begin{table}
  \centering
  \caption{The Mean Person score of different modality features compare with former works on the Hume-Reaction validation set.}
  \scalebox{0.9}{
  \begin{tabular}{@{}lcc@{}}
    \toprule
    Feature & Modality & P \\
    \midrule
    DeepSpectrum(Baseline\cite{kollias2023abaw})  & A & 0.1087 \\
    DeepSpectrum(TEMMA\cite{li2022hybrid})  & A & 0.1835 \\
    
    DeepSpectrum(Ours) & A & \textbf{0.2299}  \\
    
    VGGface2(Baseline\cite{kollias2023abaw})   & V & 0.2488 \\
    Resnet-18(TEMMA\cite{li2022hybrid})& V & 0.3893 \\
    Former-DFER+MLGCN\cite{wang2022emotional} & V & 0.3454\\
    ViPER\cite{vaiani2022viper} & V & 0.2978\\
    FaceRNET\cite{kollias2023facernet} & V & 0.3590\\
    PosterV2+Vit(Ours)& V & \textbf{0.3924} \\
    
    Baseline\cite{kollias2023abaw}  & A+V & 0.2382 \\
    ViPER\cite{vaiani2022viper}  & A+V & 0.3025 \\
    DeepSpectrum+Resnet-18(TEMMA\cite{li2022hybrid}) & A+V & 0.3968 \\
    DeepSpectrum + PosterV2-Vit(Ours) & A+V & \textbf{0.4377}\\
    DeepSpectrum + PosterV2-Vit(Ours) & A+V+AV &  \textbf{0.4394}\\
    Ensemble(Ours) & A+V+AV & \textbf{0.4452}\\
    \bottomrule
  \end{tabular}
  }
  \label{tab:results1}
\end{table}

    
    
    

\begin{table}
  \centering
  \caption{Effects of our different tricks on the Hume-Reaction validation set.}
  \begin{tabular}{@{}lcccc@{}}
    \toprule
    Model   & Class loss & EMA &  L2 Penality  & P\\
    \midrule
    Ours &  \Checkmark & \Checkmark & \Checkmark & 0.4394 \\
    Ours & \XSolid  & \Checkmark & \Checkmark & 0.4306\\
    Ours & \Checkmark & \XSolid  &  \Checkmark  & 0.4281\\
    Ours  & \Checkmark & \Checkmark & \XSolid  &   0.4165\\
    \bottomrule
  \end{tabular}
  \label{tab:results}
\end{table}

\subsection{Experimental Results}
\subsubsection{Emotional Reaction Intensity Estimation}
We conducted experiments on features from different modalities and compare the result with the baseline and Li.\cite{li2022hybrid} presented in MuSe 2022, the results are shown in Table \ref{tab:results}. We found that visual modality features were highly important, with visual scores significantly higher than audio features. After fusing the features from both modalities, model performance was further improved, reaching a Mean Person score of 0.4394 on the validation set.

We further investigated the impact of the proposed tricks on the final performance. The removal of class loss resulted in a performance decrease of 0.0088. The absence of EMA led to a performance drop of 0.0113. The removal of L2 regularization also decreased model performance by 0.0229.

Table \ref{tab:results-test} shows our final results on the test set of Hume-Reaction, and our method won the championship of this competition. 
The PCC score on the Emotional Reaction Intensity (ERI) Estimation Challenge is significantly higher than the runner-up by 0.0354.
\begin{table}
  \centering
  \caption{The final results on the Hume-Reaction test set. Numbers are provided by the web of Challenge Organization Committee.}
  \begin{tabular}{@{}lcccc@{}}
    \toprule
    Team   & PCC Score (Combined) \\
    \midrule
    \textbf{Ours} &  \textbf{0.4734} \\
    USTC-IAT-United & 0.4380\\
    Netease Fuxi Virtual Human & 0.4046 \\
    SituTech  & 0.3935 \\
    CASIA-NLPR  & 0.3865 \\
    USTC-AC  & 0.3730 \\
    NISL-2023  & 0.3667 \\
    HFUT-MAC  & 0.2527 \\
    IXLAB  & 0.1789 \\
    \bottomrule
  \end{tabular}
  \label{tab:results-test}
\end{table}

\subsubsection{Expression Classification}
\begin{table}
  \centering
  \caption{The Mean F1 score of different pre-trained methods on the Aff-Wild2 validation set. Ensemble indicates that all three pre-trained models are averaged.}
  \begin{tabular}{@{}cc@{}}
    \toprule
    Method  & F1 score \\
    \midrule
    
    Baseline(VGGface2)   & 0.23\\
    Regnet\cite{5phan2022facial}  & 0.3035 \\
    InceptionResnetV1-1(Ours)  & 0.339 \\
     InceptionResnetV1-2(Ours) & 0.284\\
     Ensemble(Ours) &  \textbf{0.346}\\
    \bottomrule
  \end{tabular}
  \label{tab:expr}
\end{table}

\begin{table}
  \centering
  \caption{The final results on the test set of Expression Classification Challenge. Numbers are provided by the web of Challenge Organization Committee.}
  \begin{tabular}{@{}lcccc@{}}
    \toprule
    Team   & F1 Score  \\
    \midrule
    Netease Fuxi Virtual Human &  0.4121 \\
    SituTech & 0.4072\\
    CtyunAI & 0.3532 \\
    HFUT-MAC  & 0.3337 \\
    HSE-NN-SberAI  & 0.3292 \\
    AlphaAff  & 0.3218 \\
    USTC-IAT-United  & 0.3075 \\
    SSSIHL DMACS  & 0.3047 \\
    DGU-IPL  & 0.2278 \\
    \textbf{HFUT-CVers (Ours)}  & \textbf{0.2277} \\
    baseline & 0.2050 \\
    \bottomrule
  \end{tabular}
  \label{tab:expr-test}
\end{table}

For comparison with the baseline model, we conducted experiments our approach in the section \ref{sec:exprcls} and further investigated the performance of three single pre-trained models.



As shown in Table \ref{tab:expr}, InceptionResnetV1-1 denotes pre-trained InceptionResnetV1 on the vggface2 dataset, InceptionResnetV1-2 denotes pre-trained InceptionResnetV1 on the casia-webface dataset. The experiment results indicated that the single pre-trained Regnet\cite{5phan2022facial} and baseline model achieved F1score of 0.3035 and 0.23 on validation set of the Aff-Wild2 database, respectively. However, our approach achieves an F1 score of 0.346, which is significantly better than the performance of the baseline model.

\section{Conclusion}
In this paper, we introduced our proposed approach to solving both Expression Classification and Emotional Reaction Intensity Estimation Challenges in ABAW5. For the Expression Classification Challenge, our approach achieves 0.346 for the F1 score on the validation set of Aff-Wild2 database. For the Emotional Reaction Intensity Estimation challenge, our model shows excellent results with a Pearson coefficient on the validation set that exceeds the baseline method by 83\%, obtaining a Mean Person score of 0.4394 on the validation set.  Furthermore, we found that the use of class loss, implementation of EMA, incorporation of regularization, and utilization of gradient clipping can greatly improve model performance.
During the ABAW5 competition, we won the first place in the Emotional Reaction Intensity Estimation track.

{\small
\bibliographystyle{ieee_fullname}
\bibliography{egbib}
}

\end{document}